# Leveraging Genetic Algorithms for Efficient Demonstration Generation in Real-World Reinforcement Learning Environments

Tom Maus (✉), Asma Atamna and Tobias Glasmachers

Ruhr-University Bochum, Bochum, Germany

`{tom.maus,asma.atamna,tobias.glasmachers}@ini.rub.de`

**Abstract.** Reinforcement Learning (RL) has demonstrated significant potential in certain real-world industrial applications, yet its broader deployment remains limited by inherent challenges such as sample inefficiency and unstable learning dynamics. This study investigates the utilization of Genetic Algorithms (GAs) as a mechanism for improving RL performance in an industrially inspired sorting environment. We propose a novel approach in which GA-generated expert demonstrations are used to enhance policy learning. These demonstrations are incorporated into a Deep Q-Network (DQN) replay buffer for experience-based learning and utilized as warm-start trajectories for Proximal Policy Optimization (PPO) agents to accelerate training convergence. Our experiments compare standard RL training with rule-based heuristics, brute-force optimization, and demonstration data, revealing that GA-derived demonstrations significantly improve RL performance. Notably, PPO agents initialized with GA-generated data achieved superior cumulative rewards, highlighting the potential of hybrid learning paradigms, where heuristic search methods complement data-driven RL. The utilized framework is publicly available and enables further research into adaptive RL strategies for real-world applications.

## 1 Introduction

The rapid evolution of industrial processes in Industry 4.0 is transforming manufacturing, emphasizing automation, customization, and efficiency. Smart factories play a central role, integrating IoT and AI to optimize production [1]. Among these technologies, reinforcement learning (RL) has emerged as a promising approach for developing intelligent control systems capable of real-time decision-making in complex, dynamic environments. With Digital Twins, virtual replicas of physical systems, industries can now safely and cost-effectively test RL solutions in realistic simulations without risking real-world disruptions [2, 3]. However, as industrial processes become more complex, advanced RL methods are needed to tackle their inherent challenges [4].

### 1.1    Reinforcement Learning

RL trains agents to learn optimal policies by interacting with an environment, receiving rewards, and refining actions based on feedback [5]. This approach has proven effective in robotics, process control, and resource management, handling uncertainty and dynamic environments [6]. Despite its successes, RL methods often face challenges like poor sample efficiency and difficulties associated with exploration, especially in sparse-reward scenarios [7].

One effective strategy to overcome these limitations involves leveraging demonstration data of successful control strategies. Demonstrations can significantly improve learning efficiency by guiding the agent's initial exploration towards promising regions of the solution space [8]. Behavioral Cloning (BC) is one such method where an agent learns by directly mimicking expert-provided actions from demonstration data, significantly speeding up the learning process [9]. However, obtaining sufficient and high-quality demonstration data is often challenging in complex industrial environments [7].

### 1.2    Genetic Algorithms

Genetic Algorithms (GAs) represent a powerful optimization technique inspired by the process of natural evolution. GAs evolve candidate solutions, represented as individuals within a population, through an iterative process including selection, crossover, and mutation. Each candidate solution is evaluated by a fitness function, in an RL context typically defined by the cumulative reward, guiding the evolution towards optimal or near-optimal strategies [10].

We propose to use GAs for generating successful demonstration trajectories by evolving candidate solutions in a simulated industrial environment. These optimized trajectories provide valuable demonstration data that can populate the replay buffer in Deep Q-Networks (DQN) or serve as a warm start for Proximal Policy Optimization (PPO) [11, 12]. We show that integrating GA-generated demonstrations into RL frameworks can significantly enhance the agent's learning performance.

We are interested in a specific industrial process, namely waste sorting. To this end, we construct an environment by combining two previously published RL benchmark environments, SortingEnv and ContainerGym, which simulate different parts of an industrial sorting process [13, 14]. It serves as a foundation for modeling a realistic yet computationally feasible industrial control system, incorporating sequential material sorting, dynamic accuracy adjustments, and process constraints that reflect real-world industrial challenges (see Fig.1). The scope of this work includes:

- presenting a new RL environment setup for benchmarking an industrial sorting process, made from a combination of two existing, complementing benchmarks,
- developing a GA-based method to generate optimal or near-optimal trajectories in the simulated industrial environment,
- integrating these trajectories into the training process of RL agents,
- evaluating the performance improvements of RL agents trained with GA-generated demonstrations compared to those trained without such demonstrations.

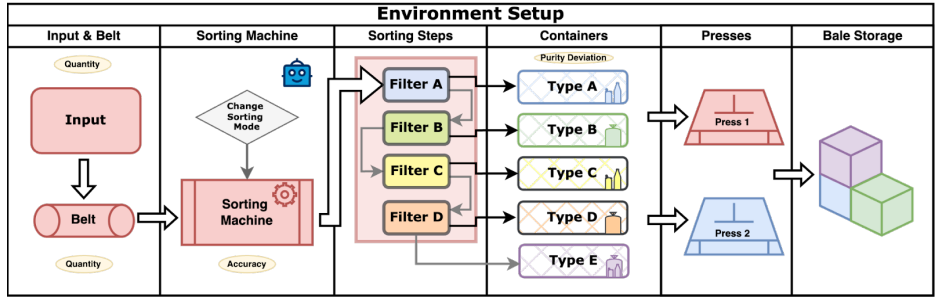

**Fig. 1.** Illustration of the sorting process, highlighting the key compartments and the interaction of the RL agent. The input station introduces mixed materials into the system, where their total quantity is observed. These materials move along the conveyor belt, where the current load is monitored before reaching the sorting machine. The sorting machine classifies materials based on their properties, with the RL agent actively adjusting the sorting mode to optimize accuracy. Once sorted, materials are collected in containers, where deviations from a defined purity threshold are tracked. The sorted materials are then transferred to the bale presses, where they are compacted into bales before being stored in the bale storage.

## 2 Related Research

### 2.1 Reinforcement Learning in Industrial and Sorting Applications

RL has emerged as a promising tool for industrial automation, particularly in process optimization and manufacturing control [6]. Unlike traditional control systems such as PID controllers or rule-based heuristics, RL enables adaptive decision-making by learning from experience. Examples are given by Lee et al., showcasing the advantages of RL-based control in nuclear power plant operations, where an RL agent outperformed conventional PID-based control systems in shutdown maneuvering [15]. Other studies have demonstrated RL's effectiveness in sorting and recycling tasks. For instance, Louette et al. applied Deep RL (TD3, SAC, PPO) to a Pick-and-Throw sorting task for scrap metal [16].

Despite promising results, RL in industrial applications faces some key challenges, particularly sample inefficiency. Training RL models requires extensive interactions, which can be costly or unsafe in real-world industrial environments. Another challenge is the lack of interpretability and safety guarantees, making it difficult for industrial engineers to trust and deploy RL-based control strategies [17]. To overcome these limitations, researchers have explored techniques such as reward shaping, curriculum learning, and domain adaptation to accelerate RL training [6, 17]. The integration of RL with expert demonstrations, as discussed in the next section, is another approach to mitigate these issues by leveraging prior knowledge for more sample-efficient training [8].

## 2.2     Expert Demonstrations and Hybrid RL Approaches

Incorporating expert demonstrations into RL training has been shown to improve learning efficiency and policy stability. One fundamental approach is Behavior Cloning (BC), which trains policies using supervised learning on expert demonstrations [18]. However, BC suffers from the distribution shift problem: If the agent deviates from the expert's trajectory, it lacks a corrective mechanism, leading to compounding errors. To address this, interactive approaches like DAgger (Dataset Aggregation) allow experts to intervene and correct agent behavior iteratively [19].

More recent methods integrate demonstrations into RL frameworks to combine the advantages of imitation learning and RL. Deep Q-learning from Demonstrations (DQfD) by Hester et al. extends Deep Q-Networks (DQN) by adding a supervised loss for expert actions, leading to faster convergence and better performance compared to standard DQN [8]. Similarly, Vecerik et al. proposed Deep Deterministic Policy Gradient from Demonstrations (DDPGfD) for continuous control, where human demonstrations are incorporated into the replay buffer to guide the RL agent's exploration. Their approach was particularly effective in a robotic task, where standard RL methods failed due to sparse rewards [20]. To improve sample efficiency, multiple techniques have been developed, such as the Prioritized Experience Replay buffer, which prioritizes important transitions based on their learning potential [21].

Overall, leveraging expert demonstrations enables RL agents to reduce exploration time and improve sample efficiency [7, 8, 20]. However, a major challenge is acquiring high-quality demonstrations, as human demonstrations can be biased or suboptimal due to factors such as distractions, limited environmental observability, or task complexity. Additionally, collecting large-scale expert demonstrations is often impractical due to the significant time and effort required [22]. This motivates research into alternative demonstration generation techniques, such as those based on Genetic Algorithms, discussed in the next section.

## 2.3     Genetic Algorithms for Generating Demonstration Data

Planners can generate optimal trajectories. However, brute-force exploration of all possible action sequences is infeasible due to the exponential growth of possibilities in long-horizon tasks. Search and optimization heuristics like GAs offer an efficient alternative by optimizing action sequences through selection, mutation, and recombination [10]. This enables structured exploration and identification of high-performing trajectories that can serve as demonstration data for RL agents.

Current research explores the use of GAs in demonstration learning. Zheng et al. introduced Genetic Imitation Learning (GenIL), where genetic operations refine existing expert demonstrations to improve reward extrapolation and policy performance [23]. Another approach explored the integration of RL with GAs for combinatorial optimization in the context of solving the traveling salesman problem, where RL-generated solutions were further refined through evolutionary search [24].

Recently, Altman et al. proposed REACT, which improves RL interpretability by using Genetic Algorithms (GAs) to select initial states that generate diverse action trajectories, eliciting edge-case behaviors and revealing policy weaknesses [25]. However, unlike our approach, REACT uses GAs to analyze policy behavior rather than optimizing trajectories for learning.

## 3    Environment and Problem Formulation

The environment utilized in this study is adapted from two published benchmarking environments (ContainerGym, SortingEnv), which address real-world industrial sorting scenarios [13, 14]. We here introduce a different learning task by conceptually combining both environments into one and adapting the sequential redistribution process, the action and observation space, reward function and underlying dynamics (e.g. of input and accuracy), to better represent the complexities in an actual industrial sorting setup.

This environment was created using the Gymnasium framework in version 0.29.1 [26] and simulates a sorting facility with a bale press, designed to process four distinct types of recyclable materials, labeled A, B, C, and D, using a set of five containers (A, B, C, D, E). Containers A-D collect materials correctly identified through a sequential sorting process, while the fifth container (E) aggregates materials that could not be sorted accurately or redistributed effectively.

The primary goal in this environment is to maintain the purity of collected recyclable materials above predefined thresholds, thus mimicking quality standards typical of real-world recycling operations. Achieving these purity targets involves managing the sorting accuracy dynamically to effectively distribute materials.

### 3.1    General Overview of Processes

The environment follows a structured sequence of operations that emulate a real-world sorting and pressing system [13, 14]. Each episode begins with the generation of a batch of input materials, progresses through the sorting process, and concludes with the compression of materials into bales for storage. The sequential process is described in the following.

1. **Generation of Input Materials:** The environment generates a batch of recyclable materials (A, B, C, and D) using a stochastic input model. This model reflects seasonal fluctuations, ensuring variability in the input material composition [14]. The generated materials serve as the initial state of the sorting process and define the complexity of the upcoming task.
2. **Agent Decision on Sorting Mode:** Before materials enter the sorting process, the agent selects the sensor setting for that batch. This binary decision determines whether the sorting accuracy is enhanced for materials A and C (mode 0) or for materials B and D (mode 1). The choice of mode influences sorting efficiency and affects the downstream purity of collected materials. This is the main mechanism for the agent to achieve the prescribed product quality goals.

3. **Materials Placed on a Conveyor Belt:** The generated batch of materials is placed onto a conveyor belt, which moves them sequentially to the sorting machine.
4. **Sequential Sorting by the Sorting Machine:** The sorting process occurs in a stepwise manner, where materials are processed sequentially at designated stations corresponding to materials A, B, C, and D. At each station, the sorting machine attempts to separate the corresponding material from the mixture on the belt. The accuracy of this separation is probabilistic and depends on the sensor setting chosen by the agent, the baseline sorting accuracy, and the belt's operational load.
5. **Classification of Sorted Materials and Residual Redistribution:** At each sorting station, correctly identified materials are deposited into their designated containers (A-D). Misclassified materials are not immediately discarded but instead redistributed to subsequent sorting stations, following an iterative redistribution mechanism. If unclassified materials remain after all sorting stations have processed them, they are collected in Container E, which acts as a final repository for unclassified residuals.
6. **Container Management and Pressing:** Each container has a defined capacity. When the fill level of any container reaches the predefined pressing threshold (e.g. 200 units), a pressing operation is triggered. Two presses are available to compact the sorted materials into bales [13]. The selection of which press to use depends on its availability, ensuring that processing is not delayed.
7. **Storage of Pressed Materials as Bales:** Once pressing is completed, materials are stored as bales. Each bale is recorded with its material type, the achieved purity level and its size, reflecting the sorting efficiency. This final step completes the operational cycle, after which the environment resets for the next episode.

This structured sequence of operations ensures that the environment captures realistic sorting and pressing challenges while maintaining a well-defined decision-making framework for the RL agent. The sequential dependencies between sorting accuracy, material redistribution, and pressing operations introduce a rich learning problem where short-term decisions affect long-term efficiency.

### 3.2   Action and Observation Space

The environment provides a structured interaction framework where the RL agent's binary action space directly influences sorting accuracy and material separation quality. At each timestep, the agent selects between two sensor settings:

- **Action 0**: Increases sorting accuracy for materials A and C.
- **Action 1**: Increases sorting accuracy for materials B and D.

The applied accuracy boost ensures near-perfect sorting (100% minus a predefined noise factor) for the selected materials, while unboosted materials remain at a baseline accuracy of 80%. Additionally, sorting accuracy declines non-linearly with increasing belt occupancy, following a squared relationship with the relative input load. This means that higher material loads lead to disproportionately larger reductions in accuracy, reflecting real-world efficiency constraints.

The observation space consists of a 33-dimensional continuous vector, encoding information on input material composition, conveyor belt state, current sorting accuracies, and material purity levels in all containers. It includes real-time purity deviations for primary containers (A–D), providing a critical learning signal directly related to the quality goals defined in the next section. By integrating both immediate state information and historical effects, the agent must anticipate long-term consequences of its sorting decisions, making the learning task highly strategic.

### 3.3   Reward Structure

The reward structure guides the RL agent to optimize sorting efficiency while maintaining purity levels above predefined thresholds. Rewards are based on purity deviations in the four primary containers, where thresholds are set at 85% for A, 80% for B, 75% for C, and 70% for D, reflecting typical quality standards in waste sorting. A positive reward is given when purity exceeds these thresholds, while a penalty is applied when it falls below them. To strongly discourage contamination, negative deviations are penalized five times more than positive deviations are rewarded.

By structuring the reward function this way, the environment strongly incentivizes purity maintenance, penalizing poor sorting outcomes disproportionately to promote strategic decision-making. Figure 2 illustrates an example of reward calculation for different threshold deviations. Since this is the sole optimization mechanism, the RL agent must adapt sensor selection to fluctuating input material amounts and compositions and current container filling status, balancing immediate sorting accuracy with long-term container purity while accounting for stochastic variations to maintain high sorting quality.

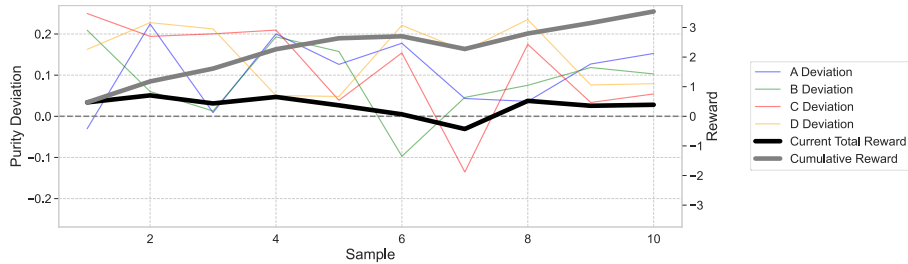

**Fig. 2.** Relationship between sorting reward and purity deviation across 10 samples. The colored lines represent the purity deviation for materials A, B, C, and D, measured as the difference between actual and threshold purity levels. The total sorting reward (black line) is derived from these deviations. The cumulative reward (gray line) demonstrates the effect of purity deviations on long-term performance.

### 3.4   Problem Formulation

The sorting environment is a sequential decision-making problem, where the RL agent dynamically adjusts sorting accuracy at each timestep. The decision for a sensor mode impacts sorting purity with a delay (time taken by the conveyor belt transport), influencing container fill levels and pressing operations. The agent must balance short-term sorting accuracy with long-term system stability, ensuring stable material throughput, minimizing contamination, and preventing inefficient pressing cycles. These complexities make the environment a rigorous testbed for RL algorithms, requiring agents to adapt to dynamic and stochastic operational constraints while optimizing overall performance. Key challenges include:

- **Cascading effects of sorting decisions**: Unlike simple RL tasks with immediate action consequences, sorting errors accumulate over time, altering container fill levels and triggering pressing events.
- **Pressing-induced state transitions**: Pressing events introduce abrupt state resets outside of the agent's control, demanding robust adaptation to dynamic conditions.
- **Stochastic variability**: Sorting accuracy depends not only on sensor mode selection but also on the belt occupancy, which reduces accuracy non-linearly.
- **Trade-off between immediate accuracy and system stability**: Misclassifications can be strategically beneficial, preventing more severe deviations in the future.
- **Long-term planning requirements**: Negative purity deviations are strongly penalized, requiring long-term anticipation rather than short-term exploitation.

## 4    Generating Demonstration Data

This section introduces our approach to generating expert demonstration trajectories using planning techniques that explore possible action sequences without relying on sequential decision-making. While these methods do not serve as viable control strategies, they can identify high-reward trajectories that provide strong reference data for RL training. We compare brute-force search and GAs to generate and evaluate candidate trajectories, which are then used as demonstrations to improve reinforcement learning performance.

### 4.1    Brute-Forcing and Genetic Algorithms for Reward Estimation

To analyze the highest achievable cumulative reward within the environment, we employed exhaustive brute-force search and evolutionary algorithms. The brute-force search explored all possible sequences of actions over a limited number of timesteps, identifying the one yielding the best possible cumulative reward. Due to combinatorial explosion, this search was limited to short action sequences (n=15).

To efficiently approximate high-reward action sequences, we implemented a GA tailored to the sorting environment. The GA evolves binary action sequences to maximize cumulative reward, using a fitness function directly based on total reward. Starting with a population of 100 randomly generated binary action sequences of given length, it applies tournament selection, where the fitter of two randomly chosen candidates is retained. Crossover occurs with a 0.7 probability, performing single-point re-combination. Mutation is applied at a per-bit rate of 0.1, meaning each action in a sequence has a 10% chance of flipping independently, introducing variability to maintain genetic diversity and prevent premature convergence. Over a given number of generations, the GA progressively refines action sequences, with the best cumulative reward determining the most effective strategy. The left plot in Fig. 3 illustrates this optimization process, highlighting the improvement in best, average, and worst-performing sequences. The highest-performing sequence serves as an upper bound for achievable policy performance, providing a benchmark for RL agents.

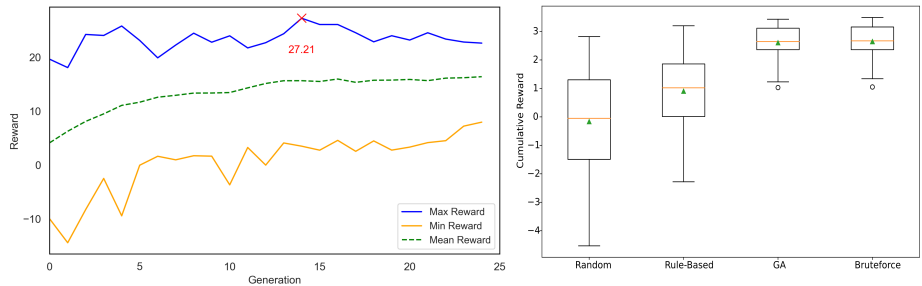

**Fig. 3.** Reward Evolution and Strategy Comparison in the Genetic Algorithm. Left: Evolution in the population over 25 generations. The lines represent the maximum reward found in each generation (blue), the minimum reward (orange), and the mean reward (green). The red marker highlights the highest observed reward. The GA was executed with a sequence length and population size of 100, running for 25 generations with a crossover rate of 0.7 and a mutation rate of 0.1 Right: Cumulative rewards across different agent strategies over 15-step action sequences. The boxplots show the reward distributions across 50 random seeds. The median reward is indicated by the orange line, while the green triangles represent the mean reward.

A comparison of the highest cumulative rewards achieved by the GA and the bruteforce solution for a shorter action sequence indicated that both methods achieved comparable performance, significantly surpassing the baseline policies, as shown in Fig. 3, right side. Building on these results, the GA was employed to generate expert demonstrations over longer sequences (n=100), capturing high-reward action trajectories. These demonstrations were stored as state-action transitions, forming a dataset for supervised pretraining and enhanced RL training, enabling experiments with imitation learning and hybrid RL approaches.

More than 240 uniquely seeded environments were used for data collection, ensuring no overlap with test environments. Each environment generated 100 state-action transitions, resulting in a diverse dataset of expert demonstrations. To maintain high-quality demonstrations, the cumulative reward of each generated trajectory was continuously compared against the performance of the rule-based agent. Only trajectories that achieved at least 15% higher cumulative reward than the rule-based solution were included in the final demonstration set. The selection results for a sample of 100 demonstrations are illustrated in Fig. 4, showing how GA-generated solutions consistently outperform both the rule-based and random policies.

It must be made clear that the trajectory generation process is not applicable as a control strategy because it has access to oracle information. In particular, the random input material composition is frozen for the length of the sequence, and in that sense it is available to the GA by entering the fitness function. This ability to "look into the future" is not available to the actual controller. It is possible only in simulation, but very useful for generating demonstrations. Therefore, planning performance is only an upper bound on the achievable performance.

### 4.2 Behavioral Cloning and Replay Buffer

To leverage expert demonstrations, behavioral cloning (BC) was implemented as a supervised learning approach using the Imitation library (v1.0.1) [27]. The BC model was trained on high-reward trajectories generated by the GA, learning a policy that maps observations to expert-selected actions. Training was performed for 100 epochs with a batch size of 256, using a fixed random seed for reproducibility. This BC-trained policy served as an initialization for the PPO model. Additionally, expert demonstrations were integrated into DQN's experience replay buffer to improve sample efficiency. The transitions were stored in a structured replay buffer, allowing DQN to benefit from both self-generated experience and expert guidance. We thus evaluated whether pretraining RL models with expert demonstrations improved learning efficiency in these two hybrid approaches:

1. **DQN with Replay Buffer (DQNRB):** The standard DQN model was trained with an experience replay buffer preloaded with expert demonstrations. This approach allowed the agent to learn from both its own exploration and high-quality expert examples, potentially accelerating convergence.
2. **PPO with Warm Start (PPOBC):** The BC-trained model was used to initialize the neural network weights of a PPO agent. This warm-starting strategy provided a structured initial policy, leveraging supervised learning for early-stage policy shaping while allowing PPO to refine strategies through RL.

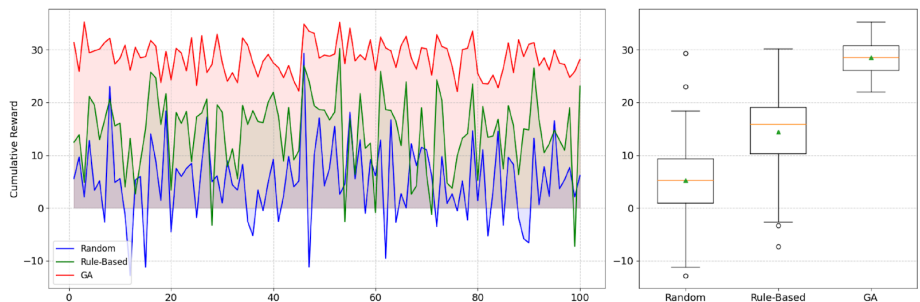

**Fig. 4.** Cumulative rewards of different agent strategies for 100 differently seeded environment simulations with each 100 steps. Left: Cumulative reward per seed for the Random (blue), Rule-Based (green), and GA (red) agents. Shaded areas indicate the reward range observed across multiple runs. Right: Distribution of cumulative rewards per agent showing median (orange line) and mean (green triangle) rewards.

## 5     Experiments and Results

To evaluate the performance of different RL strategies within the sorting environment, we conducted a series of experiments, including baselines with random and rule-based agents, RL model training, and heuristic optimization using brute-force search and GAs. A key research question guiding our experiments was to what extent GA-generated demonstration trajectories improve RL performance.

### 5.1     Baseline Performance: Random and Rule-Based Agents

Before training RL models, we established two baselines: a random agent and a rule-based agent. The random agent selects sensor settings randomly at each timestep, providing an unstructured baseline that represents the expected performance of an uninformed policy. The rule-based agent, in contrast, follows a simple decision heuristic that prioritizes sensor settings based on the relative proportions of materials on the conveyor belt. Specifically, if the sum of materials A and C exceeds the sum of B and D, the agent selects the mode that boosts A and C, otherwise favoring B and D. These baselines serve as reference points for evaluating RL performance.

### 5.2     Reinforcement Learning Agents

We trained two common deep RL models, Deep Q-Networks (DQN) and Proximal Policy Optimization (PPO), utilizing the implementation from Stable-Baselines3 v2.2.1 [28]. DQN utilizes experience replay and neural network approximations to learn optimal value functions, whereas PPO is a policy gradient method widely recognized for stability and efficiency in continuous control tasks [11, 12].

Both models were trained for one million timesteps using default hyperparameters from Stable-Baselines3, except for explicitly defined settings in the following. PPO was configured with an entropy coefficient of 0.01 to encourage exploration and used a multi-layer perceptron (MLP) policy with two hidden layers of 32 neurons each. DQN was trained with experience replay but without demonstration data or additional optimizations. All training runs were initialized with a fixed random seed to ensure reproducibility. Throughout training, an evaluation callback assessed model performance every 10,000 timesteps using an evaluation environment with a fixed seed. At the end of training, the fully trained model was compared against the best-performing model identified by the callback in a series of 10 evaluation episodes. The model with the highest average reward was retained as the final agent.

### 5.3     Performance Comparison of Agent Strategies

To comprehensively evaluate all tested models, we conducted a large-scale benchmarking experiment that compared the effectiveness of different approaches across multiple environments. Figure 5 summarizes their performance.

A total of 100 uniquely seeded environments were evaluated, each running for 100 steps, ensuring diverse input conditions. The tested models include random selection (R), rule-based heuristics (RB), deep RL models (DQN, PPO), hybrid RL models utilizing expert demonstrations (DQNRB, PPOBC), and genetic optimization (GA). Each model was tested under identical conditions, and the cumulative reward was recorded. RL models were loaded from pre-trained checkpoints and executed in evaluation mode. The GA was re-run for each seed to optimize action sequences within the given constraints. All tested seeds were distinct from those used during training and demonstration data collection, ensuring no overlap or data leakage.

- The random agent (R) performed the worst, reinforcing the need for structured decision-making in this task.
- The rule-based agent (RB) significantly outperformed random selection, providing a strong heuristic baseline.
- Both DQN and PPO models, trained from scratch, exceeded rule-based performance, demonstrating the effectiveness of RL in this domain.
- DQNRB did not yield a notable improvement over standard DQN, indicating limited benefits from passive demonstration integration.
- PPOBC substantially outperformed standard PPO, confirming that incorporating expert knowledge accelerates learning and improves overall performance.
- The GA achieved the highest overall reward, surpassing all other models, including RL-based approaches. However, since the GA operates as an offline optimization method, it is not suitable for real-time action selection.

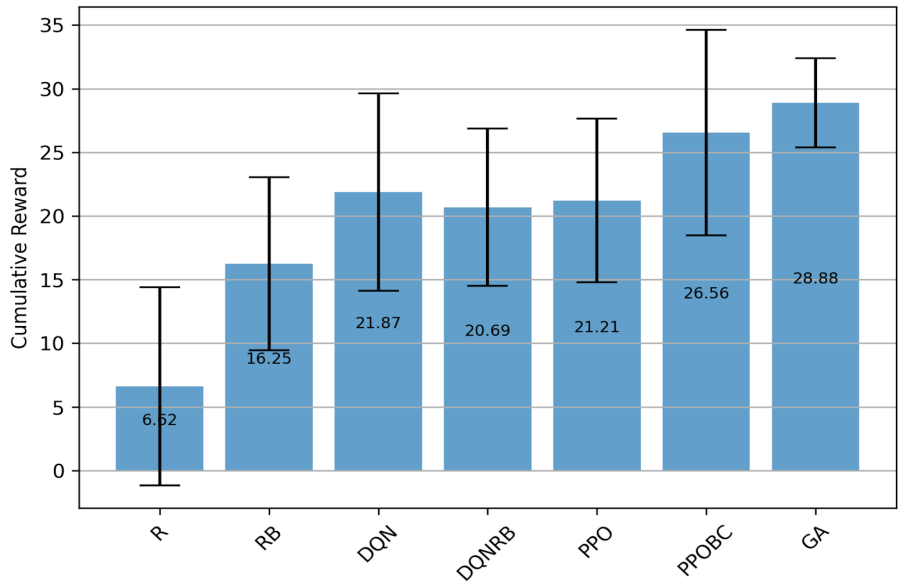

**Fig. 5.** Benchmarking results of different agent strategies over 100 random seeds. The bars represent the mean cumulative reward for each approach, with error bars indicating the standard deviation.

# 6    DISCUSSION AND CONCLUSION

Many studies show that RL benefits from expert demonstrations, improving policy convergence, sample efficiency, and training stability, particularly in complex decision-making tasks [6, 7, 9, 17, 22]. However, most RL research relies on simplified benchmarks rather than real-world industrial applications [8, 24, 25]. To address this, we introduced a sorting environment combining SortingEnv and ContainerGym [13, 14], evaluating expert demonstration generation and its impact on RL performance. The environment simulated a sequential decision-making task where an RL agent optimizes sorting accuracy while managing throughput constraints. We compared brute-force search and GA for generating expert trajectories, with GA proving more efficient in identifying high-reward sequences. These GA-derived demonstrations were used to train Deep Q-Networks (DQN) and Proximal Policy Optimization (PPO), incorporated via replay buffer augmentation for DQN (DQNRB) and behavioral cloning pretraining for PPO (PPOBC).

We found that GA-based optimization provided a strong upper bound for achievable performance and GA-generated demonstrations significantly improved PPO training (PPOBC), leading to faster convergence and higher cumulative rewards compared to standard PPO. DQN with a replay buffer of expert trajectories (DQNRB) did not yield notable improvements, indicating that simple replay augmentation may not be sufficient. The GA achieved the highest cumulative rewards, demonstrating its effectiveness in identifying near-optimal sorting strategies. However, it is an offline optimization method using privileged information and unsuitable for real-time decision-making. Thus, for real-world deployment, PPOBC emerged as the best-performing model, balancing adaptability, efficiency, and robustness in live industrial environments.

Our study provides a practical demonstration of how heuristic optimization can enhance RL training, particularly in industrial automation. Many real-world RL applications struggle with data scarcity and inefficient exploration, and our results show that GA-generated data can serve as an alternative to human-labeled demonstration data. By making both the environment and training framework publicly available, we enable future research to build upon our work, fostering further advancements in RL for real-world decision-making tasks. In the following, we want to highlight some limitations and areas for further research:

- Scalability and Computational Feasibility: Our studied environment benefitted from GA-based demonstrations, but scaling to larger action spaces remains challenging, as exhaustive testing becomes computationally infeasible. [17, 22]. Future research should explore adaptive GAs or hybrid search strategies to improve scalability.
- Replay Buffer Limitations in DQN: Limited improvements in DQNRB suggest that basic replay buffer augmentation is insufficient. To better utilize expert demonstrations, strategies like prioritized experience replay or DQfD should be tested [8, 21].
- Parameter Tuning and Real-World Constraints: Unlike digital twins with predefined ranges from operational data, artificially designed environments require manual tuning [3]. This risks bias in defining feasible parameter spaces. Future work should explore data-driven parameter selection or domain adaptation to enhance realism.

To conclude, this study showed that GAs can generate expert demonstration data that can be used to significantly enhance RL training, achieving faster convergence and higher cumulative rewards compared to training from scratch. This highlights the potential of hybrid heuristic-RL approaches for industrial automation. Future work should assess the feasibility of scaling GA-based demonstrations to complex RL tasks and explore alternative ways to leverage expert data in value-based RL methods.

**Acknowledgement**: This research received external funding from the German Federal Ministry for Economic Affairs and Climate Action through the grant "EnSort".

**Code Availability:** The code for blind review can be found underline{here.}

**Disclosure of Interests.** The authors have no competing interests to declare.